\documentclass[10pt,twocolumn,letterpaper]{article}

\usepackage{cvpr}
\usepackage{times}
\usepackage{epsfig}
\usepackage{graphicx}
\usepackage{amsmath}
\usepackage{amssymb}

\usepackage{lipsum}
\usepackage[sort]{cite}
\usepackage{multirow}

\usepackage[breaklinks=true,bookmarks=false]{hyperref}

\newcommand{\Xp}{{S_q^+}}

\newcommand{\ndp}{h_{k}^+}

\newcommand{\nbits}{b}
\newcommand{\sgn}{\mathop{\mathrm{sgn}}}

\newcommand{\ours}{DOAP}

\cvprfinalcopy 

\def\cvprPaperID{368} 

\begin{document}

\title{Local Descriptors Optimized for Average Precision}

\author{Kun He\\
Boston University\\
{\tt\normalsize hekun@bu.edu}
\and
Yan Lu\thanks{Now with Nvidia.}\\
Honda Research Institute USA\\
{\tt\normalsize sinoluyan@gmail.com}
\and 
Stan Sclaroff\\
Boston University\\
{\tt\normalsize sclaroff@bu.edu}
}

\maketitle

\begin{abstract}
Extraction of local feature descriptors is a vital stage in the solution pipelines for numerous computer vision tasks. Learning-based approaches improve performance in certain tasks, but still cannot replace handcrafted features in general. In this paper, we improve the learning of local feature descriptors by optimizing the performance of descriptor matching, which is a common stage that follows descriptor extraction in local feature based pipelines, and can be formulated as nearest neighbor retrieval. Specifically, we directly optimize a ranking-based retrieval performance metric, Average Precision, using deep neural networks. This general-purpose solution can also be viewed as a listwise learning to rank approach, which is advantageous compared to recent local ranking approaches. 
On standard benchmarks, descriptors learned with our formulation achieve state-of-the-art results in patch verification, patch retrieval, and image matching.
\end{abstract}


\section{Introduction}
\label{sec:intro}

Extracting feature descriptors from local image patches is a common stage in many computer vision tasks involving alignment or matching.
To replace handcrafted feature engineering, recently much attention has been paid to learning local feature descriptors.
Despite exciting progress, certain levels of handcrafting are currently present in the design of learning objectives for local feature descriptors, making it difficult to have performance guarantees when the learned descriptors are integrated into larger pipelines. 
Indeed, according to a recent study \cite{comparative}, traditional handcrafted features such as SIFT \cite{SIFT} can still outperform learned ones in complicated tasks such as 3D reconstruction.
In this paper, we aim to improve the learning of local feature descriptors by optimizing better objective functions.

Our thesis is that local feature descriptor learning is not a standalone problem, but rather a component in the optimization of larger pipelines. Therefore, the learning objectives should be designed in accordance with other pipeline components.
Upon inspection of common local feature matching pipelines, we find that feature matching can be exactly formulated as nearest neighbor retrieval.
Thus, we propose a novel listwise \emph{learning to rank} formulation for learning local feature descriptors,
based on the direct optimization of a ranking-based retrieval performance metric: Average Precision.
Our formulation uses deep neural networks, and works for both binary and real-valued descriptors. 
Compared to recent approaches, our method optimizes a commonly adopted evaluation metric, and eliminates complex optimization heuristics.
Descriptors learned with our formulation achieve state-of-the-art results in benchmarks including UBC Phototour \cite{DAISY}, HPatches \cite{HPatches}, RomePatches \cite{RomePatches}, and the Oxford dataset \cite{Mikolajczyk2005}.

An important feature of our proposed formulation is that it is general-purpose, as it optimizes the performance of the task-independent nearest neighbor matching stage,
rather than a task-specific pipeline.
Nevertheless, to better tailor the learned descriptors for feature matching, we also augment our formulation with task-specific improvements.
First, we make use of the Spatial Transformer module \cite{STN} to effectively handle geometric noise and improve the robustness of matching, without requesting extra supervision.
Also, for the challenging HPatches dataset, we design a clustering-based technique to mine additional patch-level supervision, which improves the performance of learned descriptors in the image matching task.

In summary, we propose a general-purpose learning to rank formulation that optimizes local feature descriptors for nearest neighbor matching.
Our learned descriptors achieve state-of-the-art performance, and are further enhanced by task-specific improvements.
We believe that our contribution can serve as a stepping stone for the direct optimization of larger computer vision pipelines.

\begin{figure*}[ht]
\centering
\includegraphics[width=.83\linewidth]{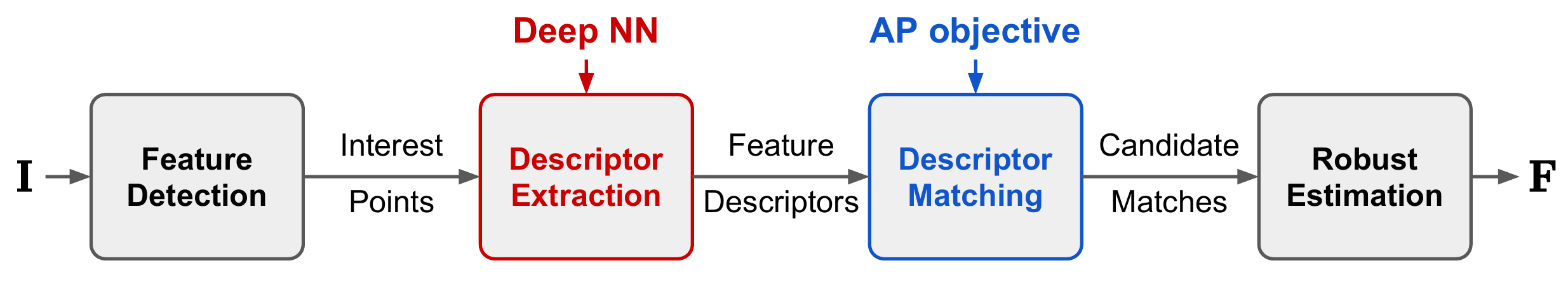}
\caption{An example local feature-based image matching pipeline, where the task is to estimate the fundamental matrix $\mathbf{F}$ between images $\mathbf{I}=(I_1,I_2)$, using robust estimation techniques such as RANSAC \cite{RANSAC}.
We model the {feature descriptor extractor} using {deep neural networks}, and directly optimize {a ranking-based objective (Average Precision)} for the subsequent stage of descriptor matching.
}
\label{fig:pipeline}
\end{figure*}

\section{Related Work}
\label{sec:related}

\noindent\textbf{Learning Local Features}

Parallel with the long history of handcrafted computer vision pipelines (the most prominent example being SIFT \cite{SIFT}), numerous researchers have attempted to replace handcrafted components with learned counterparts. 
There exist many formulations for learning different components in local feature based pipelines.
For example, interest point detectors are learned in \cite{TILDE,covariant,QuadNets}, LIFT \cite{LIFT} learns three components separately in a feature matching pipeline, and DSAC \cite{DSAC} approximately learns a camera localization pipeline end-to-end.

For learning local feature descriptors, some early works use simple architectures \cite{DAISY,BinBoost} and convex optimization \cite{Simonyan14}.
Later approaches use deep neural networks: PhilippNet \cite{PhilippNet} learns by fitting pseudo-classes, DeepDesc \cite{DeepDesc} applies Siamese networks, MatchNet \cite{MatchNet} and DeepCompare \cite{DeepCompare} learn nonlinear distance metrics for matching, and \cite{RomePatches} uses Convolutional Kernel Networks.
A series of recent works have considered more advanced model architectures and triplet-based deep metric learning formulations, including UCN \cite{UCN}, TFeat \cite{TFeat}, GLoss \cite{Gloss}, L2Net \cite{L2Net}, HardNet \cite{HardNet}, and GOR \cite{GOR}.

Instead of optimizing  triplet-based surrogate losses,  we employ  listwise learning to rank to directly optimize the performance of the matching stage.
Although end-to-end optimization of the pipeline is attractive, it is unfortunately highly difficult and task-dependent.
By focusing on the two task-independent stages (descriptor extraction and matching), our solution is general-purpose and can be potentially integrated into larger optimization pipelines.

\vspace{.2em}
\noindent\textbf{Evaluating Local Feature Descriptors}

Local features ideally should be evaluated in terms of final task performance, \eg Mikolajczyk and Schmid  \cite{Mikolajczyk2005} use precision and recall derived from image matching, and Schonberger \etal \cite{comparative} use a benchmark based on 3D reconstruction.
However, in complex vision pipelines, final task performance can be affected by individual components. 
For example, \cite{HPatches} observes that without controlling for components such as interest point detection in image-based benchmarks,  different conclusions can be drawn when comparing the relative performance of feature descriptors.

Patch-based benchmarks provide  unambiguous evaluation for local feature descriptors.
The \emph{patch verification} task is first proposed in \cite{DAISY}, formulated as binary classification on the relationship between patch pairs.
RomePatches \cite{RomePatches} and HPatches \cite{HPatches} both consider the \emph{patch retrieval} task, which simulates nearest neighbor matching, and is shown \cite{HPatches} to be more realistic and challenging compared to patch verification.
A ranking-based evaluation metric, Average Precision, is adopted in both benchmarks.

\vspace{.2em}
\noindent\textbf{Ranking Optimization in Metric Learning}

Metric learning \cite{metric_survey} is a general family of methods that learn distance functions from data.
While much previous effort focused on learning Mahalanobis distances, recently the metric learning community has focused on learning vector embeddings to be used with standard (\eg Euclidean) distance metrics.
In this light, the problem of learning local feature descriptors is an instance of metric learning.

Learning vector embeddings  necessarily calls for task-dependent formulations. 
For nearest neighbor retrieval,  optimization of ranking performance has been studied in metric learning. 
For example, learning to rank formulations for Mahalanobis distances are proposed in \cite{MLR,Lim2014}. 
Triplet-based deep metric learning approaches \cite{Law2017ICML,histloss,LiftStruct,samplingmatters} can also be viewed as optimizing surrogate ranking losses.
In the ``learning to hash'' subcommunity that considers the special case of learning binary embeddings, He \etal \cite{TALR} directly optimize ranking-based retrieval performance measures with deep neural networks, based on an approximation to histogram binning originally proposed in \cite{histloss}, which is also adopted in learning binary descriptors by \cite{mihash}.
We make use of their optimization technique in the learning of binary and real-valued descriptors for our problem.


\section{Optimizing Descriptors for Matching}
\label{sec:method}

In this section, we motivate our approach by analyzing the descriptor matching stage, and point out that it corresponds to nearest neighbor retrieval.
Then we discuss a learning to rank formulation to optimize ranking-based retrieval performance. 

\subsection{Nearest Neighbor Matching}
\label{sec:method:nn}

Consider Fig.~\ref{fig:pipeline}, which depicts a pipeline for estimating the fundamental matrix between matching images $I_1$ and $I_2$.
It consists of four stages: feature detection, descriptor extraction, descriptor matching, and robust estimation.
Suppose we detect and extract $M$ local features from each image.
The descriptor matching stage operates as follows:
it computes the pairwise distance matrix with $M^2$ entries,
 and for each feature in $I_1$, looks for its nearest neighbor in $I_2$, and vice versa. 
Feature pairs that are mutual nearest neighbors\footnote{For simplicity, the distance ratio check \cite{SIFT} is not considered.} become candidate matches in the robust estimation stage, such as RANSAC \cite{RANSAC}.

We point out that this matching process is exactly performing nearest neighbor retrieval: each feature in $I_1$ is used to query a database, which is the set of features in $I_2$. 
For good performance, true matches should be returned as top retrievals, while false matches are ranked as low as possible.
Performance of the matching stage also directly reflects the quality of the learned descriptors, since it has no learnable parameters (only performs distance computation and sorting).
To assess nearest neighbor matching performance, we adopt Average Precision (AP), a commonly used evaluation metric.
AP evaluates the performance of retrieval systems under the \emph{binary relevance} assumption: retrievals are either ``relevant'' or ``irrelevant'' to the query.
This naturally fits the local feature matching setup, where given a reference feature, features in a target image are either its true match or false match.
Next, we learn binary and real-valued local feature descriptors to optimize AP. 

\subsection{Optimizing Average Precision}
\label{sec:method:AP}

We first introduce mathematical notation.
Let $\mathcal{X}$ be the space of image patches, and $S\subset\mathcal{X}$ be a database.
For a query patch $q\in\mathcal{X}$, let $S_q^+$ be the set of its matching patches in $S$, and let $S_q^-$ be the set of non-matching patches.
Given a distance metric $D$, let $(x_1,x_2,\ldots,x_n)$ be a ranking of items in $S_q^+\cup S_q^-$ sorted by increasing distance to $q$, \ie $D(x_1,q)\leq D(x_2,q) \ldots \leq D(x_n,q)$.
Given the ranking, AP is the average of precision values ($Prec@K$) evaluated at different positions:
\vspace{-.5em}
\begin{align}
Prec@K&=\frac{1}{K}\sum_{i=1}^K \boldsymbol{1}[x_i\in S_q^+],\\
AP & = \frac{1}{|S_q^+|}\sum_{K=1}^n \boldsymbol{1}[x_K\in S_q^+]Prec@K,
\end{align}
where $\boldsymbol{1}[\cdot]$ is the binary indicator. 
AP achieves its optimal value if and only if every patch from $S_q^+$ is ranked above all patches from $S_q^-$.

The optimization of AP can be cast as a metric learning problem, where the goal is to learn a distance metric $D$ that gives optimal AP when used for retrieval.
Ideally, if all the above steps can be formulated in differentiable forms, then AP can be optimized by exploiting chain rule. 
However, this is not possible in general: the sorting operation, required in producing the ranking, is non-differentiable, and continuous changes in the input distances induce discontinuous ``jumps" in the value of AP.
Thus, appropriate smoothing is necessary to derive differentiable approximations of AP.

Our solution is based on a recent result in the metric learning community.
For the problem of learning binary image-level descriptors for image retrieval, He \etal \cite{TALR} observe that sorting on integer-valued Hamming distances can be implemented as histogram binning, 
and employ a differentiable approximation to histogram binning \cite{histloss} to optimize ranking-based objectives with gradient descent.
We use this optimization framework to optimize AP for both binary and real-valued local feature descriptors.
In the latter case, the optimization is enabled by a novel quantization-based approximation that we develop.

\vspace{.3em}
\noindent\textbf{Binary Descriptors}

Binary descriptors offer compact storage and fast matching, which are useful in applications with speed or storage restrictions.
Although binary descriptors can be learned one bit at a time \cite{BinBoost}, here we take a gradient-based relaxation approach  to learn fixed-length ``hash codes".

Formally, a deep neural network $F$ is used to model a mapping from patches to a low-dimensional Hamming space: ${F}\!:\!\mathcal{X}\rightarrow\{-1,1\}^b$. 
For the Hamming distance $D$, which takes integer values in $\{0,1,\ldots,b\}$, AP can be computed in closed form using entries of a histogram $\mathbf{h}^+=(h_0^+,\ldots,h_b^+)$, where $h_k^+=\!\sum_{x\in\Xp}\boldsymbol{1}[D(q,x)\!=\!k]$.
The closed-form AP can further be continuously relaxed, and differentiated with respect to $\mathbf{h}^+$ \cite{TALR}.

The next step in the chain rule is to differentiate entries of $\mathbf{h}^+$ with respect to the network $F$. 
Usnitova and Lempitsky \cite{histloss} approximate the histogram binning operation as 

\vspace{-1em}
\begin{align}
h_k^+ \approx \sum_{x\in\Xp}\delta(D(q,x),k),
\label{eq:histbin}
\end{align}

\vspace{-.2em}
\noindent replacing the binary indicator with a differentiable function $\delta$ that peaks when $D(q,x)=k$.
This allows to derive approximate gradients as
\vspace{-.2em}
\begin{align}
\frac{\partial \ndp}{\partial F(q)} & \approx \sum_{x\in \Xp}
\frac{\partial \delta(D(q,x),k)}{\partial D(q,x)}
\frac{\partial D(q,x)}{\partial F(q)},
\label{eq:chainrule1}\\
\frac{\partial \ndp}{\partial F(x)} & \approx \boldsymbol{1}[x\in\Xp]
\frac{\partial \delta(D(q,x),k)}{\partial D(q,x)}
\frac{\partial D(q,x)}{\partial F(x)}.
\label{eq:chainrule2}
\end{align}
Note that the partial derivative of the Hamming distance is obtained via this differentiable formulation:
\begin{align}
D(x,x')  = \frac{1}{2}\left(\nbits-F(x)^\top F(x')\right). \label{eq:hamming_dist}
\end{align}

Finally, the thresholding operation used to produce binary bits is smoothed using the $\tanh$ function,
\vspace{-.2em}
\begin{align}
{F}(x) & = (\sgn(f_1(x)),\ldots,\sgn(f_b(x))) \\
& \approx (\tanh(f_1(x)),\ldots,\tanh(f_b(x))), \label{eq:sigphi}
\end{align}

\vspace{-.2em}
\noindent where $f_i$ are real-valued neural network activations.
With these relaxations, the network can be trained end-to-end.

\vspace{.3em}
\noindent\textbf{Real-Valued Descriptors}

To complete our formulation, we next consider real-valued descriptors, which are preferred in high-precision scenarios.
We model the the descriptor as a vector of real-valued network activations, and apply $L_2$ normalization: $\|F(x)\|=1,\forall x$. 
In this case, the Euclidean distance $D$ is given as
\vspace{-.6em}
\begin{align}
D(x,x') =\sqrt{2-2F(x)^\top F(x')}.
\label{eq:Euclidean}
\end{align}

\vspace{-.3em}
The main challenge in optimizing AP for real-valued descriptors is again the non-differentiable sorting, but  real-valued sorting has no simple alternative form.
However, histogram binning can be used as an approximation: we \emph{quantize} real-valued distances using histogram binning, obtain the histograms $\mathbf{h}^+$, {and then reduce the optimization problem to the previous one.}
With $L_2$-normalized vectors, the quantization is easy to implement as the Euclidean distance has closed range $[0,2]$: we simply uniformly divide $[0,2]$ into $b+1$ bins.
To derive the chain rules in  this case, only the partial derivatives of the distance function needs modification in \eqref{eq:chainrule1} and \eqref{eq:chainrule2}.
The differentiation rules for the $L_2$ normalization operation are well known, and we give full derivations in the appendix.

Differently from the case of binary descriptors, the number of histogram bins $b$ is now a free parameter, which involves a trade-off. On the one hand, a large $b$ reduces quantization error, which in fact achieves zero if each histogram bin contains at most one item. 
On the other hand, gradient computation for approximate histogram binning has linear complexity in $b$. Nevertheless, in our experiments, we consistently obtain good results using $b\leq 25$.

\subsection{Comparison with Other Ranking Approaches}
\label{sec:method:LTR}

We would like to contrast our approach with others in the learning to rank context. 
Some recent methods, \eg \cite{L2Net,HardNet,TFeat,GOR}, learn feature descriptors by optimizing losses defined on triplets in the form of $(a,p^+,p^-)$, where $a$ is an anchor patch, $p^+$ is its matching patch, and $p^-$ is a non-matching patch. 
The loss typically encourages the learned distance metric $D$ to satisfy $D(a,p^+)<D(a,p^-)-\rho$, where $\rho$ is a margin.
Triplet losses have a long history in metric learning \cite{triplet2004,triplet2010}, and are better suited for ranking tasks than pair-based losses used in Siamese networks (\eg \cite{DeepDesc}).
In learning to rank terminology \cite{learn_rank}, triplets define local \emph{pairwise} ranking losses, 
while our approach is \emph{listwise} since the evaluation metric  that we optimize (AP) is defined on a ranked list.

\begin{figure}[t]
\centering
\includegraphics[width=\linewidth]{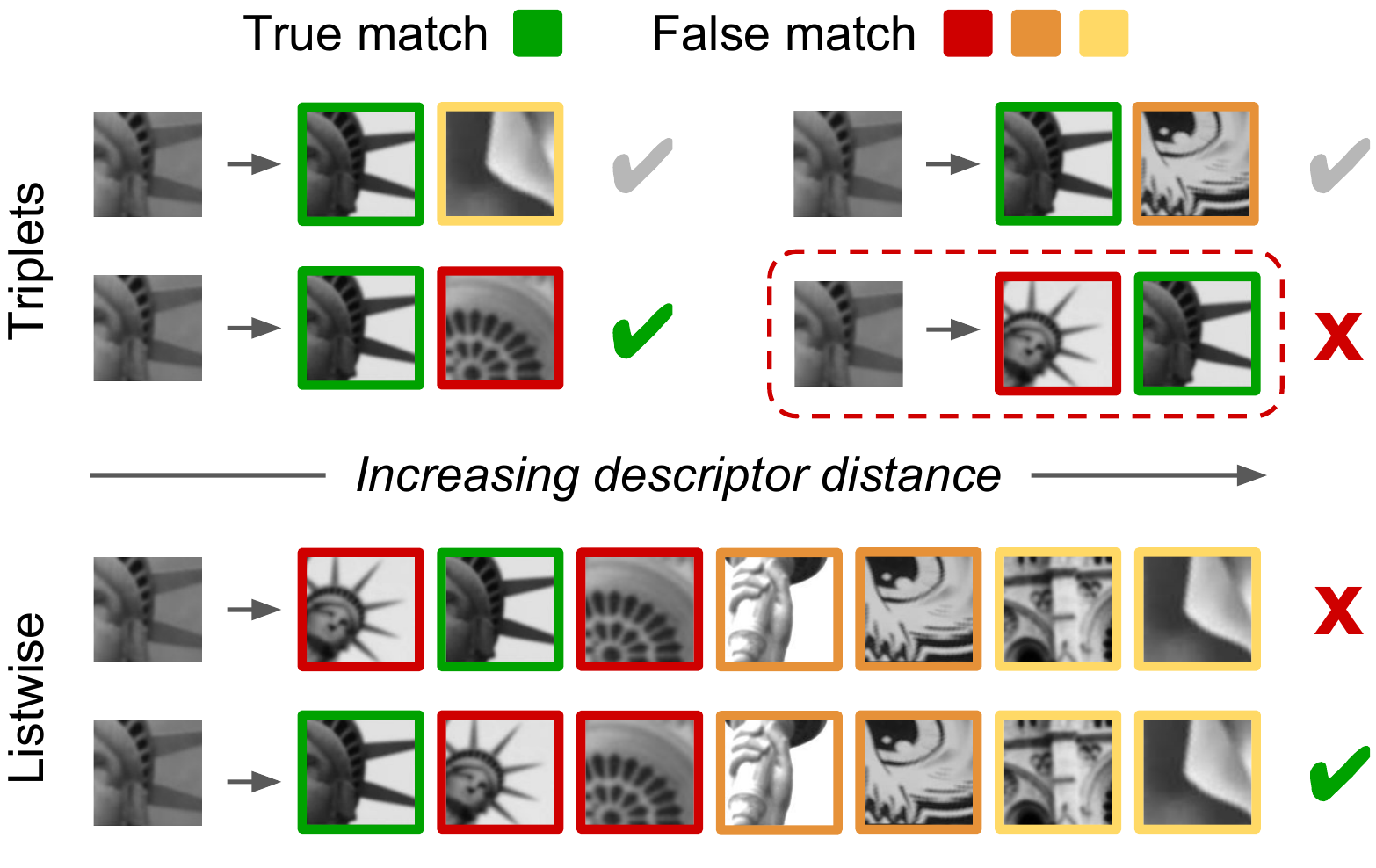}
\caption{Comparison between triplet-based and listwise ranking approaches.
Top: in triplet-based training, most triplets get correctly classified early (first row), and it is crucial to find and correct high-rank errors (red dashed box), with a heuristic known as hard negative mining.
Bottom: in listwise ranking which is \emph{position-sensitive}, the high-rank error would reduce AP from $1$ to $0.5$, thus automatically receiving a heavy penalty.
Our listwise optimization corrects such errors  without using complex mining heuristics.
Best viewed in color.
}
\label{fig:AP}
\end{figure}

Despite their simplicity, triplet losses can be very challenging to optimize.
For $N$ training examples, the set of triplets is of size $O(N^3)$,  but most of them get classified correctly early on during learning.
To maintain stable progress, carefully tuned heuristics such as hard negative mining \cite{HardNet}, anchor swap \cite{TFeat}, or distance-weighted sampling \cite{samplingmatters} are crucial.
We note that these optimization difficulties stem from a fundamental mismatch between triplet losses and listwise evaluation.
As shown in Fig.~\ref{fig:AP}, listwise metrics are \emph{position-sensitive}, while local losses are insensitive; an error made on a single triplet may have a big impact on the result if it occurs near the top of the list.
Therefore, heuristics are needed to focus on reducing high-rank errors.
In contrast, our method directly optimizes the listwise evaluation metric, Average Precision, and is free of such heuristics.
The listwise optimization also implicitly encodes hard negative mining:   it requires matching patches to be ranked above all non-matching patches, which automatically enforces correct classification of the hardest triplet in the batch without explicitly finding it.

\section{Task-Specific Improvements}
\label{sec:addon}

In addition to the general-purpose learning to rank formulation, we develop two improvements that take the nature of local feature matching into account.

\subsection{Handling Geometric Noise}
\label{sec:addon:STN}
To improve the robustness of local features for matching, it is key to build invariance to  {geometric} noise into the descriptor:
SIFT \cite{SIFT} estimates orientation and affine shape to normalize input patches, and LIFT \cite{LIFT} includes a learned orientation estimation module.
Likewise, we {can} also include a geometric alignment module in our descriptor networks.
Our choice is the Spatial Transformer \cite{STN},  which aligns input patches by predicting a 6-DOF affine transformation, without requiring extra supervision.
In our experiments, this module is able to correct geometric distortion, and consistently improve performance.

In contrast to the image-based UCN \cite{UCN}, which also includes Spatial Transformers, 
our patch-based networks have limited input size, and the predicted affine transformation can often lead to out-of-boundary sampling, which corrupts sampled patches.
We address this challenge by using appropriate boundary padding. 
Details are given in the appendix.

\subsection{Label Mining for Image Matching}
\label{sec:addon:kmeans}
While our formulation directly optimizes for the task of \emph{patch retrieval}, 
it is also possible to address higher-level tasks.
We demonstrate this with the \emph{image matching} task in the challenging HPatches dataset \cite{HPatches}, which contains patches extracted from matching image sequences.

{The image matching task in HPatches is formulated similarly as patch retrieval, which involves retrieving matching patches in a pool of ``distractors".
However, the distractors are defined differently.
In patch retrieval, distractors do not include patches in the same image sequence as the query, due to concern of repeating structures in images.
In image matching, images are matched against others in the {same} sequence, which means that all distractors are actually in-sequence.
Thus, image matching performance can be improved by including in-sequence distractors when optimizing patch retrieval.

We perform \emph{label mining} to augment the set of distractors when optimizing patch retrieval in HPatches.  To avoid noisy labels in the presence of repeating structures, we use a simple heuristic: clustering. 
For each image sequence, we cluster all patches based on visual appearance. Then, patches having high inter-cluster distance are marked as distractors for each other (with 3D verification).
Note that label mining is not related to the hard negative mining heuristic, since its goal is to add additional supervision.
Please see Sec.~\ref{sec:exp:HPatches} and appendix for more details.

\section{Experiments}
\label{sec:exp}

\begin{figure}
\centering
\includegraphics[width=\linewidth]{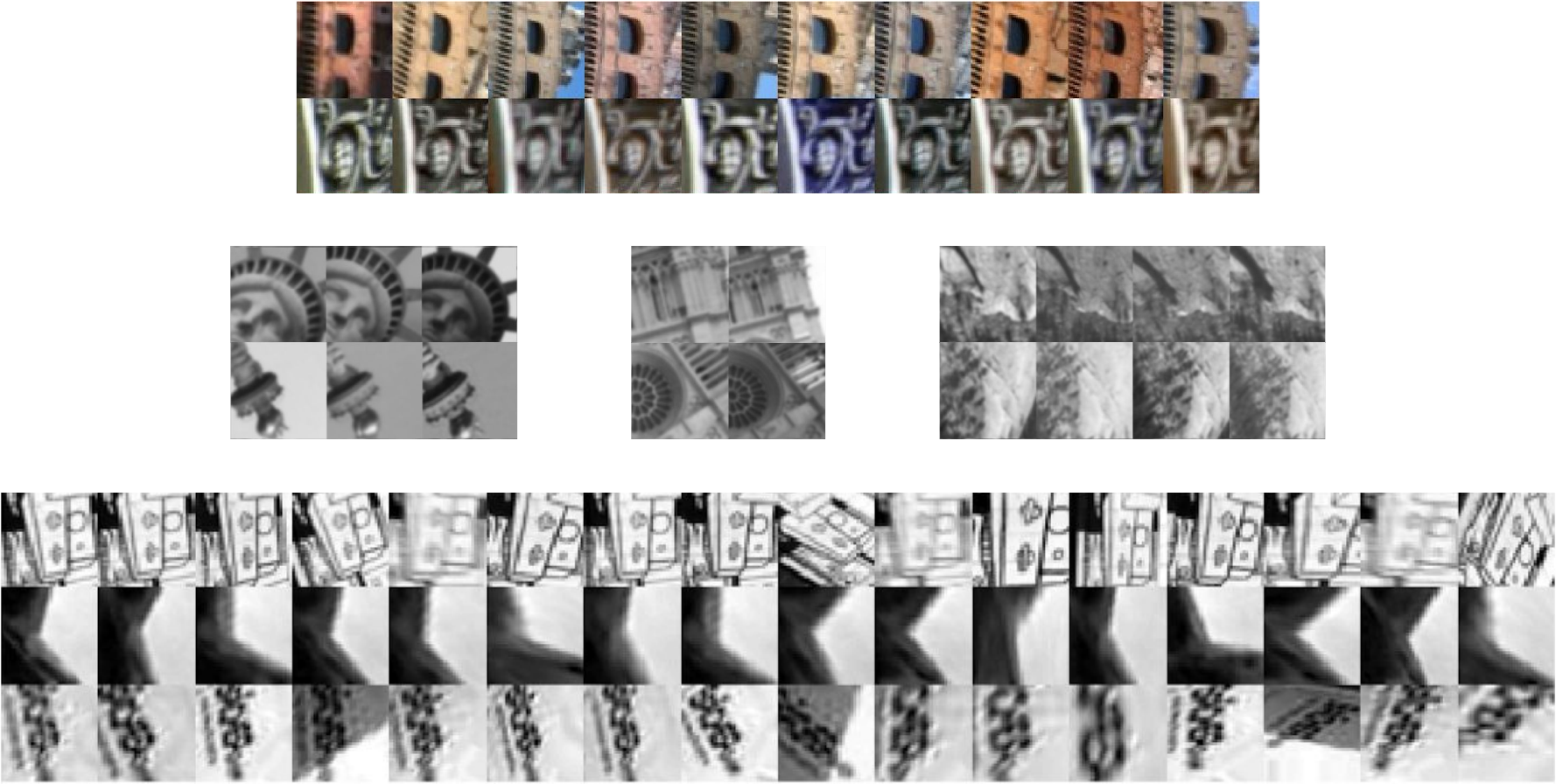}
\caption{Examples from three patch-based datasets (top to bottom): RomePatches \cite{RomePatches}, UBC Phototour \cite{DAISY}, HPatches \cite{HPatches}.  
In all datasets, patches are grouped such that patches in the same group correspond to the same 3D point.
}
\label{fig:patches}
\end{figure}

We experiment with three patch-based datasets (examples are in Fig.~\ref{fig:patches}): UBC Phototour \cite{DAISY}, HPatches \cite{HPatches}, and RomePatches \cite{RomePatches}.
We use the CNN architecture recently proposed in L2Net \cite{L2Net}, which consists of seven convolution layers, and is regularized with Batch Normalization and Dropout.
We do not use the more complex ``Center Surround'' architecture.
The input to the network is 32x32 grayscale, and we resize input patches to this size.
When adding the Spatial Transformer module, we increase the input size to 42x42, and use 3 convolution layers to predict a 6-DOF affine transformation, which is then used  to sample a 32x32 patch.

\begin{table*}[ht]
\centering
\begin{tabular}{c|c|cc|cc|cc|c} 
\hline
\multirow{2}{*}{\bf Method} & {Train} & Notredame & Yosemite & Liberty  & Yosemite & Liberty & Notredame  &  FPR95 \\
\cline{2-8}
& {Test} & \multicolumn{2}{c|}{Liberty} & \multicolumn{2}{c|}{Notredame} & \multicolumn{2}{c|}{Yosemite} & {Mean} \\
\hline
\multicolumn{9}{c}{\it Real-valued descriptors} \\
\hline
SIFT \cite{SIFT}  & 128
& \multicolumn{2}{c|}{29.84}  & \multicolumn{2}{c|}{22.53}   & \multicolumn{2}{c|}{27.29}  & 26.55          \\
MatchNet\cite{MatchNet}  & 128
& 7.04    & 11.47  & 3.82    &  5.65   & 11.6   & 8.70     & 8.05   \\
TFeat-M* \cite{TFeat}    & 128
& 7.39    & 10.31   & 3.06    &  3.80   & 8.06    & 7.24     & 6.64   \\
TL-AS-GOR \cite{GOR}  & 128
& 4.80  & 6.45  & 1.95  & 2.38  & 5.40  & 5.15  & 4.36 \\       
DC-2ch2st+ \cite{DeepCompare}  & 512
& 4.85  & 7.20  & 1.90  & 2.11  & 5.00  & 8.39  & 4.19 \\
CS-SNet-GLoss+ \cite{Gloss}  & 256
& 3.69  & 4.91  & 0.77  & 1.14  & 3.09  & 2.67  & 2.71 \\
L2Net+ \cite{L2Net}  & 128
& 2.36  & 4.7   & 0.72  & 1.29  & 2.57  & 1.71  & 2.23 \\
HardNet+ \cite{HardNet}   & 128
& 2.28  & 3.25  & 0.57  & 0.96  & 2.13  & 2.22  & 1.90 \\       
HardNet-GOR+ \cite{HardNet,GOR}  & 128
& 1.89  & 3.03  & 0.54  & 0.90  & 2.41  & 2.39  & 1.86 \\       
CS-L2Net+ \cite{L2Net}  & 256
& 1.71  & 3.87  & 0.56  & 1.09  & 2.07  & 1.30  & 1.76 \\       
\ours{}+ & 128
& 1.54  &  2.62 & 0.43  & 0.87  & 2.00  & \textbf{1.21}  & 1.45 \\       
\ours{}-ST+ & 128
& \textbf{1.47}  &  \textbf{2.29} & \textbf{0.39}  & \textbf{0.78}  & \textbf{1.98}  & 1.35  & \textbf{1.38} \\       
\hline
\multicolumn{9}{c}{\it Binary descriptors} \\
\hline
BinBoost \cite{BinBoost}  & 64
& 20.49  & 21.67   & 16.90  & 14.54  & 22.88  & 18.97  & 19.24 \\
L2Net+ \cite{L2Net}  & 128
& 7.44  & 10.29   & 3.81  & 4.31  & 8.81  & 7.45  & 7.01 \\
CS-L2Net+ \cite{L2Net}  & 256
& 4.01  & 6.65   & 1.90  & 2.51  & 5.61  & 4.04  & 4.12 \\
\ours{}+ & 256
& 3.18  &  4.32 & 1.04 & \textbf{1.57}  & {4.10}  & 3.87  & 3.01 \\       
\ours{}-ST+ & 256
& \textbf{2.87}  & \textbf{4.17} & \textbf{0.96}  & {1.76}  & \textbf{3.93}  & \textbf{3.64}  & \textbf{2.89} \\       
\hline
\end{tabular}
\vspace{.5em}
\caption{Patch verification performance on UBC Phototour, where metric is false positive rate at 95\% recall (FPR95). The best results are in \textbf{bold}.
Second column shows dimensionality, and methods with suffix ``+'' are trained with data augmentation.
Both the binary and real-valued versions of \ours{} and \ours{}-ST achieve state-of-the-art results.
}
\label{table:ubc}
\end{table*}

We name our {descriptor} \ours{} ({\bf D}escriptors {\bf O}ptimized for {\bf A}verage {\bf P}recision), and test its binary and real-valued versions.
Our networks are trained using SGD with momentum 0.9 and weight decay $10^{-4}$, and {the} learning rate is decayed linearly to zero within a fixed number of epochs. 
The initial learning rate (always on the order of 0.1) and number of epochs are tuned during training. 
Input normalization is as follows: patches are normalized by subtracting the  mean pixel value in the patch and then dividing by the standard deviation.

\subsection{UBC Phototour}
\label{sec:exp:UBC}

We first conduct experiments on the UBC Phototour dataset \cite{DAISY}, a classical benchmark of descriptor performance.
Patches are extracted from Difference-of-Gaussian detections in three image sequences: \emph{Liberty}, \emph{Notre Dame}, and \emph{Yosemite}.
Following the standard setup, we use six training/test combinations formed by the three sequences,
and report patch verification performance in terms of false positive rate at 95\% recall (FPR95).

We train our models on UBC Phototour with data augmentation, in the form of random flipping and 90-degree rotations, which showed consistent performance improvement in previous work.
We compare to a range of existing descriptors, including both binary and real-valued, listed in Table~\ref{table:ubc}.
L2Net \cite{L2Net} and HardNet \cite{HardNet} are two leading methods, which optimize triplet-based losses with the same CNN architecture as ours.
We also include methods that use the ``Center Surround'' architecture: CS-SNet-Gloss \cite{Gloss} and CS-L2Net,
and we have applied the recent global regularization technique in \cite{GOR} to HardNet, resulting in a more competitive method which we call HardNet-GOR.
Compared to existing approaches, \ours{} achieves state-of-the-art performance with both binary and real-valued descriptors, and results are further improved by \ours-ST, which includes the Spatial Transformer module.

We attribute the  performance of \ours{} and \ours-ST to the listwise AP optimization.
As mentioned in Sec.~\ref{sec:method:LTR}, listwise optimization automatically includes the ``hard negative mining" heuristic in local ranking approaches, since it implicitly enforces the correct classification of all induced pairs and triplets.
We then expect performance to improve when increasing training batch size, as larger batches lead to longer lists and increased likelihood of including hard negatives.
We validate this by training the 128-dimensional \ours{} model on \emph{Liberty}, varying batch size between 256 and 4096,  and monitoring the average of FPR95 on \emph{Notre Dame} and \emph{Yosemite}.
Indeed, Fig.~\ref{fig:batchsize} shows that performance improves with batch size and saturates after 2048.
Similar trends are also observed in HardNet \cite{HardNet}, with saturation occurring at batch size 512.
In contrast, the listwise optimization allows the performance of DOAP to  saturate at a later stage.

\begin{figure}
\vspace{-.5em}
\centering
\includegraphics[width=.8\linewidth]{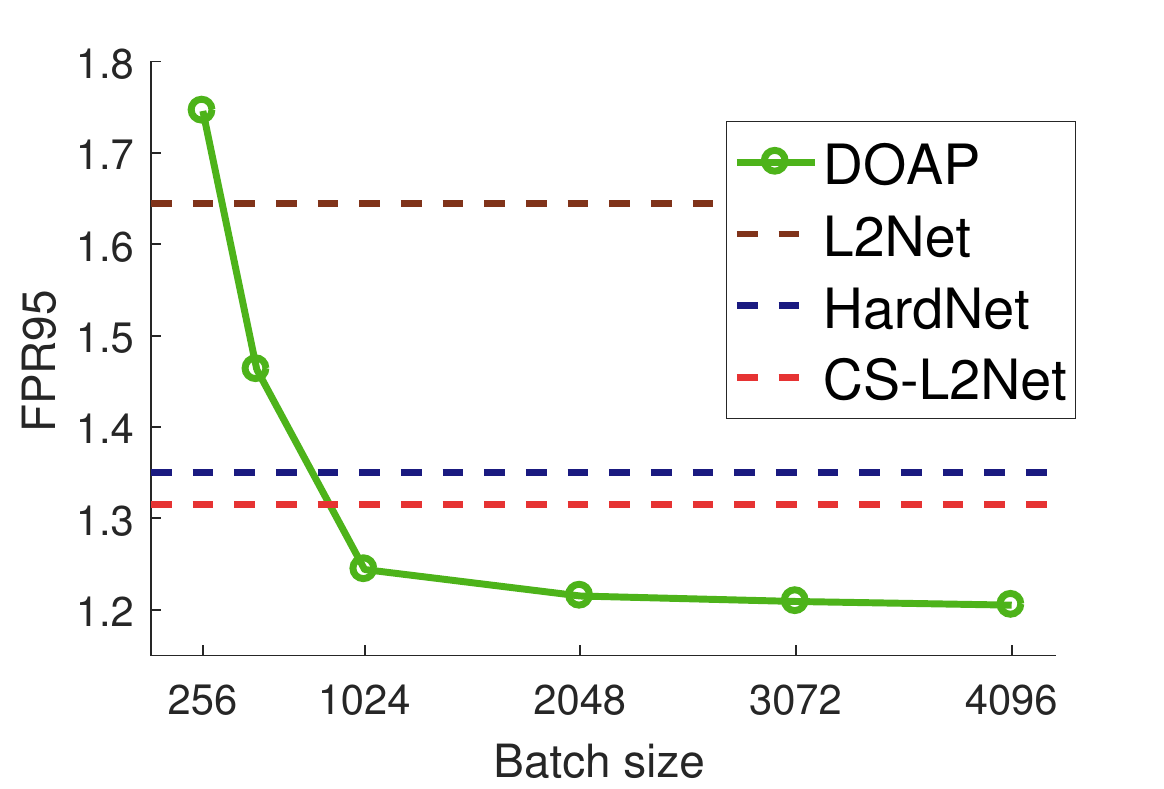}
\caption{Influence of training batch size for the 128-d \ours{} descriptor trained on \emph{Liberty}, with data augmentation. Vertical axis: average of FPR95 on \emph{Notre Dame} and \emph{Yosemite}.
}
\label{fig:batchsize}
\vspace{-.5em}
\end{figure}

\subsection{HPatches}
\label{sec:exp:HPatches}

\begin{figure*}[ht]
\centering
\includegraphics[height=1.7em]{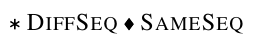}
\includegraphics[height=1.7em]{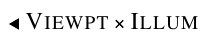}
\includegraphics[height=1.7em]{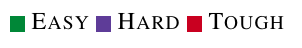}

\vspace{-1.2em}
\includegraphics[width=.33\linewidth]{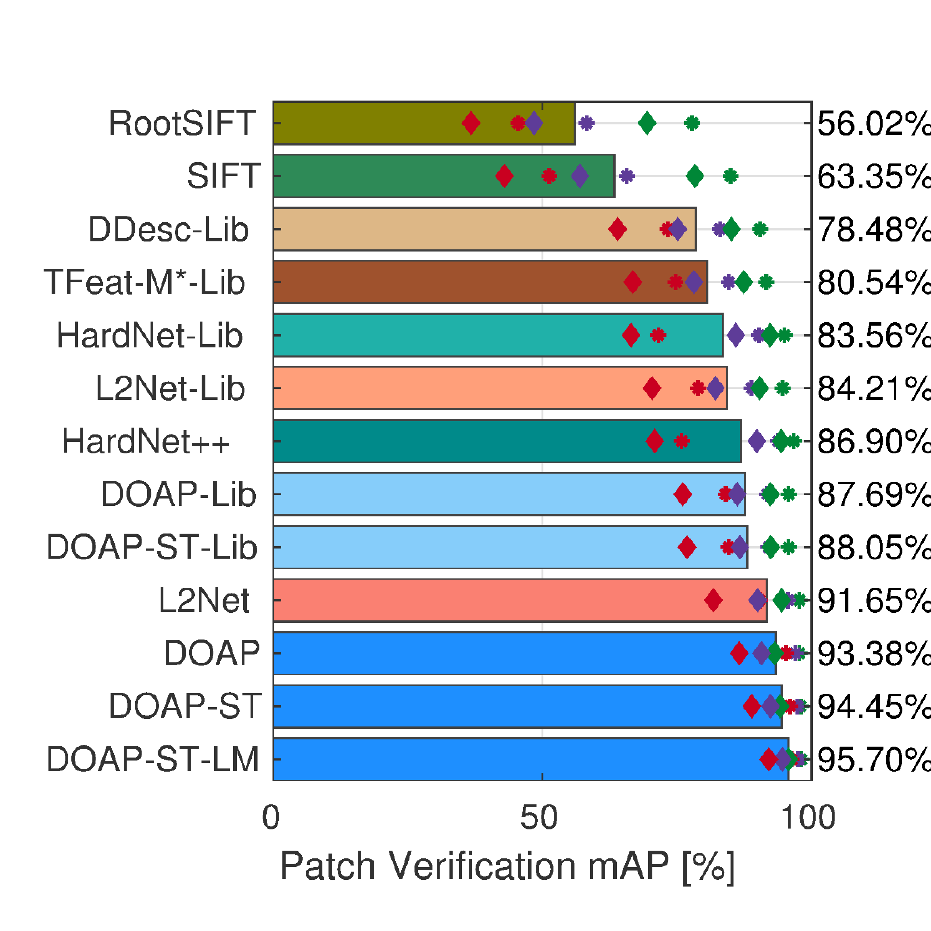}
\includegraphics[width=.33\linewidth]{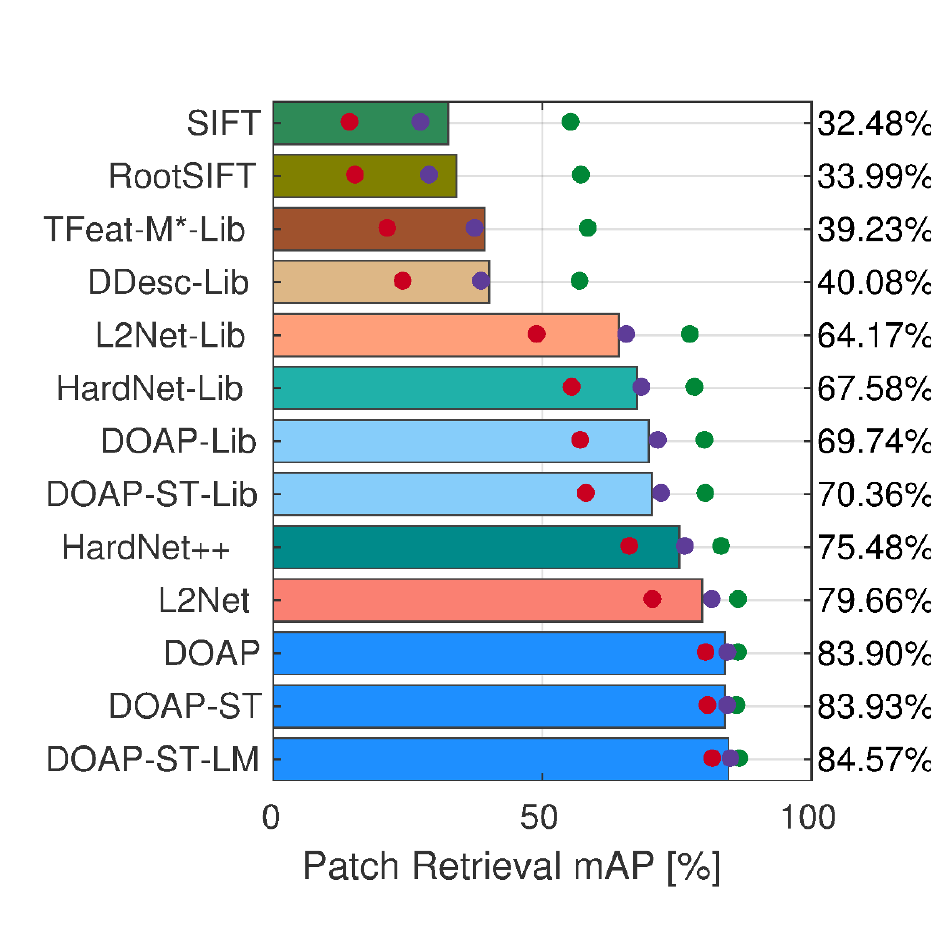}
\includegraphics[width=.33\linewidth]{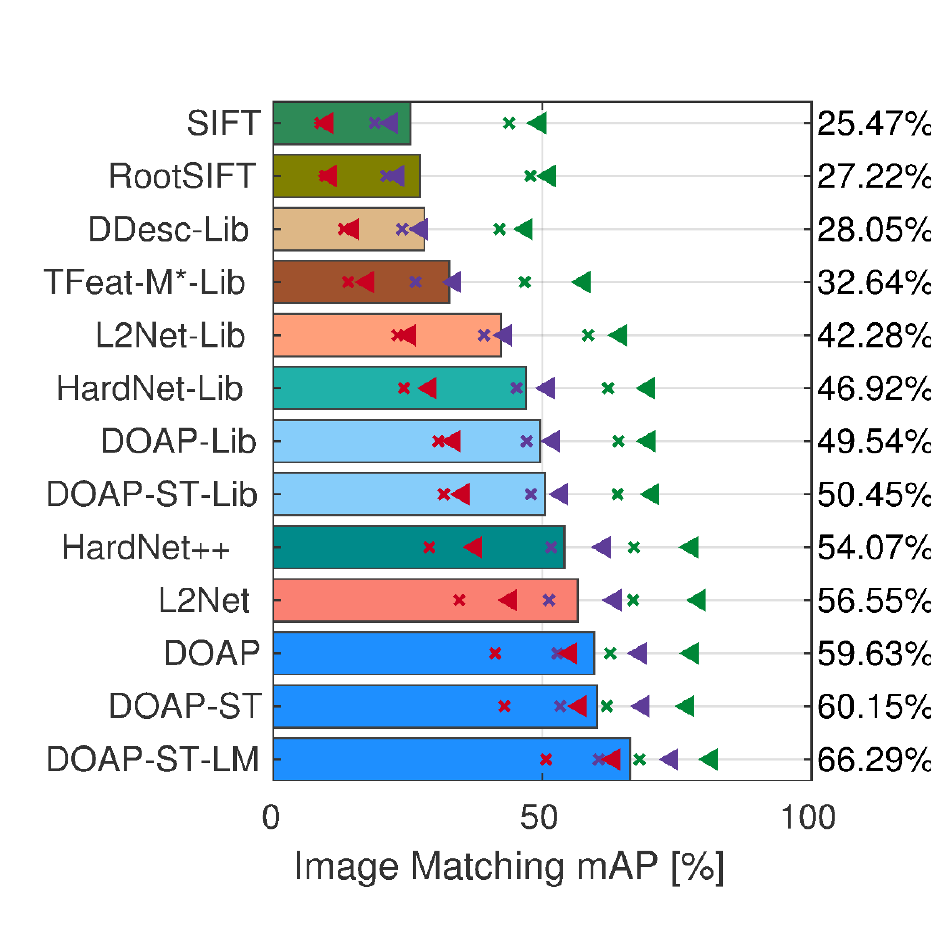}

\vspace{-.5em}
\caption{Results on the HPatches dataset, evaluated on the test set of the ``a'' split.   No ZCA normalization \cite{HPatches} is used.
Suffix indicates training set used (Lib: \emph{Liberty}, no suffix: HPatches). HardNet++ is trained on the union of \emph{Liberty} and HPatches.
\ours{} outperforms competing methods in all tasks, and all of its variants  excel in handling tough test cases. 
}
\label{fig:HPatches}
\end{figure*}

HPatches \cite{HPatches} consists of a total of over 2.5 million patches extracted from 116 image sequences, each with 6 images with known homography.
Both viewpoint and illumination changes are included, and test cases have levels of difficulty \emph{easy}, \emph{hard}, and \emph{tough}, according to the amount of geometric noise.
Three evaluation tasks are considered (in increasing order of difficulty): patch verification, patch retrieval, and image matching.

In this experiment, we focus on comparing real-valued descriptors. 
We first include four baselines reported in \cite{HPatches}: SIFT \cite{SIFT}, RootSIFT \cite{RootSIFT}, DeepDesc \cite{DeepDesc}, and TFeat \cite{TFeat}.
Next, as results for L2Net and HardNet trained on the \emph{Liberty} sequence of UBC Phototour are reported in \cite{HardNet}, for fair comparison, we also report results for our models trained on \emph{Liberty}.
Finally, we train and evaluate three versions of our descriptor on HPatches: \ours{}, \ours{}-ST with the Spatial Transformer, and \ours{}-ST-LM, which additionally uses label mining.
We compare to the L2Net model trained on HPatches,  and HardNet++, trained on the union of \emph{Liberty} and HPatches.
Note that CS-L2Net is excluded as it performs worse than L2Net in this more realistic dataset, which is consistent with the observations in \cite{L2Net,Gloss}.
When determining training/test sets, we use the ``a'' split: 
the test set contains 40 image sequences (20 viewpoint and 20 illumination), and the training set contains the other 76 sequences.

Fig.~\ref{fig:HPatches} presents results 
on HPatches.\footnote{
Results for L2Net and HardNet are obtained using their publicly released models and may slightly differ from those reported in \cite{HardNet}.
}
Our descriptors achieve state-of-the-art results for all three tasks, and all variants are better at handling \emph{tough} test cases than competing methods. 
Specifically, \ours{} and \ours{}-ST obtain the best patch retrieval performance, which  directly results from the optimization of patch retrieval mAP. 
This optimization also gives state-of-the-art performance in patch verification.
For the most challenging task of image matching, as mentioned in \cite{HPatches}, patch retrieval performance is well correlated. 
However, due to the difference in task definition that we mentioned in Sec.~\ref{sec:addon:kmeans}, all methods see lower performance when tested for image matching.
With the clustering-based label mining, \ours-ST-LM significantly improves image matching mAP compared to the next best models: around {6}\%  and {10}\% over \ours-ST and L2Net, respectively.
Notably, it achieves \textbf{over 50\% mAP} even in the toughest test cases (\emph{tough} geometric noise, illumination change).
The inclusion of extra supervision also boosts patch retrieval performance, since in-sequence distractors provide harder negatives to learn from.

\begin{table}
\centering
\begin{tabular}{c|cc|cc}
\hline
\textbf{Method} & \textbf{Coverage} & \textbf{Dim.} & \textbf{Train} & \textbf{Test}  \\
\hline
SIFT \cite{SIFT} & 51x51 & {128} & 91.6 & 87.9   \\
AlexNet-conv3 \cite{AlexNet} & 99x99 & 384 & 81.6 & 79.2  \\
PhilippNet \cite{PhilippNet} & 64x64 & 512 & 86.1 & 81.4  \\
CKN-grad \cite{RomePatches} & 51x51 & 1024 & 92.5 & 88.1  \\
\ours{} & 51x51 & {128} & \textbf{95.9} & \textbf{88.4}  \\ 
Binary \ours{} & 51x51 & {256} & 95.2 & 86.8  \\ 
\hline
\end{tabular}
\vspace{.2em}
\caption{Patch retrieval mAP comparison on RomePatches.
SIFT is a strong baseline, previously only surpassed by the high-dimensional CKN-grad \cite{RomePatches}.
DOAP is the first descriptor to outperform SIFT with the same dimensionality.
}
\label{table:rome}
\end{table}

\begin{figure*}[t]
\centering
\includegraphics[width=\linewidth]{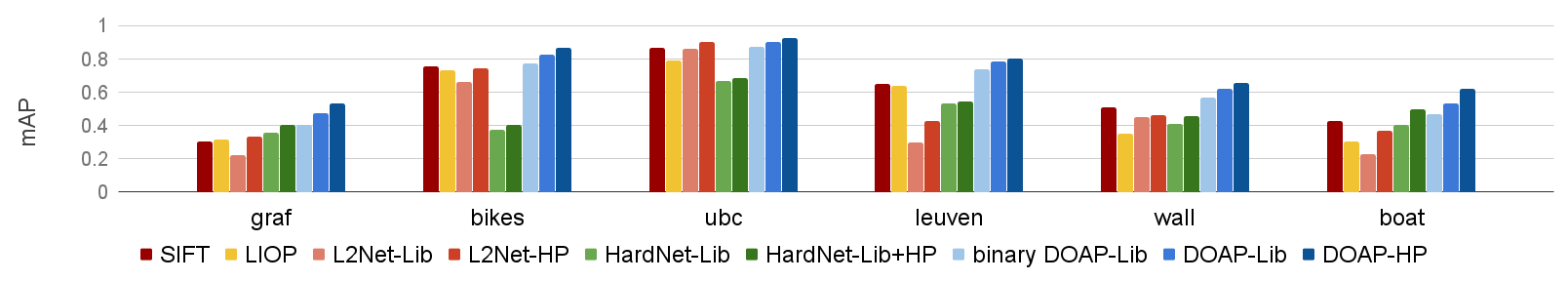}
\caption{Image matching performance on the Oxford dataset \cite{Mikolajczyk2005}. 
Suffixes indicate  the training set used (Lib: \emph{Liberty}, HP: HPatches).
Here, all versions of DOAP include the Spatial Transformer.
}
\label{fig:matching}
\end{figure*}

\subsection{RomePatches}
\label{sec:exp:Rome}
We next consider the RomePatches dataset \cite{RomePatches}, which contains 20,000 image patches of size 51x51, split equally into training and test sets.
The task is patch retrieval.
This dataset is constructed by performing SIFT matching on images taken in Rome, and keeping matching patches that satisfy 3D constraints.
With such tailored construction, SIFT is unsurprisingly a strong baseline on RomePatches.
In fact, in terms of test set mAP, previous methods, 
including pretrained AlexNet \cite{AlexNet} and PhilippNet \cite{PhilippNet}, could not surpass SIFT.
The only method to do so was the CKN-grad variant proposed in \cite{RomePatches}, using 1024-dimensional descriptors.

Due to the small size of RomePatches, 
we found it necessary to increase weight decay in SGD to $5\times 10^{-4}$, and Dropout rate from $0.1$ to $0.5$ in the L2Net architecture.
Also, adding Spatial Transformers did not improve results, possibly because the patches are already well aligned (see examples in Fig.~\ref{fig:patches}); therefore we only report results for the binary and real-valued DOAP.
As seen in Table~\ref{table:rome}, the real-valued \ours{} outperforms SIFT and other descriptors with \textbf{88.4\% mAP} on the test set, while the binary version also performs competitively.
The comparison between DOAP and SIFT is fair, since they have the same input coverage and output dimensionality.
Note that the closest competitor to DOAP, CKN-grad \cite{RomePatches}, is unsupervised and needs high dimensionality to perform well.
By exploiting supervised learning and directly optimizing the evaluation metric, we are able to get better training and test performance while using 8x fewer dimensions (128 \vs 1024).

\subsection{Image Matching in Oxford Dataset}
\label{sec:exp:retrieval}

Lastly, we use our learned descriptors to perform image matching in six image sequences from the classical Oxford dataset \cite{Mikolajczyk2005}, where the matching pipeline also includes interest point detection.
We use the implementation from \texttt{VL-Benchmarks} \cite{vlbenchmarks}; features are detected by the Harris-Affine detector, and then patches are extracted with a magnification factor of 3 relative to the detected feature frames.
The evaluation metric is mean Average Precision (mAP), computed as the area under the precision-recall curve derived from nearest neighbor matching.

We compare to SIFT, LIOP \cite{LIOP} (the best-performing handcrafted descriptor in \cite{L2Net}'s experiment), and 128-d real-valued versions of L2Net and HardNet with different training sets. 
We use the 256-bit binary and 128-d versions of DOAP trained on \emph{Liberty}, and the 128-d version trained on HPatches.
From the results in Fig.~\ref{fig:matching}, we can see that SIFT is indeed difficult to beat, and good results on the UBC benchmark does not guarantee high-level task performance, especially in the case of HardNet.
{The real-valued DOAP consistently outperforms SIFT and other descriptors with significant margins, especially in the more challenging sequences such as \emph{graf} and \emph{boat}.
The binary DOAP trained on \emph{Liberty} also outperforms other real-valued descriptors on average, including L2Net trained on HPatches, and HardNet trained on the union of \emph{Liberty} and HPatches.
}

\subsection{Discussion}

\vspace{-.3em}
\noindent\textbf{Minibatch Sampling.}
We discuss the minibatch sampling strategy used in training our models.
First, note that  in all datasets considered, patches are provided in groups: patches within a group correspond to the same 3D point and thus match each other (see Fig.~\ref{fig:patches}). 
The group size, denoted $n$, is between 2 and 3 on average in UBC Phototour, and equals 10 in RomePatches.
For HPatches, $n=16$, as each patch has a reference version, and five variations from each difficulty level.

{Our sampling strategy differs from those in local ranking approaches, where} patch groups are often broken up to form pairs or triplets in a pre-processing step before training.
Instead, we directly sample \emph{groups} to construct training minibatches, so that patches belonging to the same group are always in the same batch.
This allows our listwise optimization to utilize supervision with maximum efficiency. 
Let minibatch size be $M$, every training patch is associated with a listwise ranking constraint, that its $n-1$ matches need to be ranked at the top of a list of size $M-1$.
This constraint alone needs $(n-1)(M-n)$ triplets to fully capture.
Take UBC Phototour as an example, assuming $n=2.5$ on average, a single minibatch of size $1024$ induces about $1.6\times 10^6$ triplets, which is already $1/32$ of the total number of training triplets used in HardNet.
For HPatches ($n=16$), this number would be $1.5\times 10^7$.
However, triplets do not need to be explicitly generated in our listwise optimization.

\vspace{.1em}
\noindent\textbf{Time Complexity.}
For a minibatch of size $M$, the pairwise distances between all examples are computed, and then binned into $b$-bin histograms. The time complexity is $O(bM^2)$. 
The quadratic dependency on $M$ is in fact optimal, due to distance computation.

There is also a trade-off involving the batch size $M$.
A larger batch size leads to longer lists and better performance, but slows training.
Nevertheless, even with $M=4096$, a single training epoch on \emph{Liberty} takes less than 4 minutes on an Nvidia Titan X Pascal GPU.
Similar to the case of UBC (Fig.~\ref{fig:batchsize}), performance saturation is also observed around $M=2048$ in HPatches and RomePatches.

\vspace{-.3em}
\section{Conclusion}
\vspace{-.2em}
In this work, we use deep neural networks to learn binary and real-valued local feature descriptors that optimize nearest neighbor matching performance.
This is achieved through a listwise learning to rank formulation that directly optimizes Average Precision. Our formulation is general-purpose, and is superior to recent local ranking approaches.
We further enhance our formulation with task-specific components:  
handling geometric noise with the Spatial Transformer, and mining labels using clustering.
The learned descriptors achieve state-of-the-art performance in patch verification, patch retrieval, and image matching.
Future work will explore the optimization of larger portions in vision pipelines, for example, by incorporating differentiable versions of robust estimation.

\vspace{-.2em}
\section*{Acknowledgements}
\vspace{-.2em}
A major part of this work was done during KH's internship at Honda Research Institute.
This work is also partly conducted at Boston University, supported by a BU IGNITION award, NSF grant 1029430, and gifts from Nvidia.

{\small
\bibliographystyle{plain}
\bibliography{egbib}
}

\vfill
\pagebreak
\onecolumn

\appendix
\section*{Appendix}

\section{Learning Real-Valued Descriptors}
We model the mapping from image patches to descriptors as $F\!:\!\mathcal{X}\to\mathcal{Y}$, where $\mathcal{Y}$ is the descriptor space, and $F$ is a neural network. With real-valued descriptors, we take $\mathcal{Y}=\mathbb{R}^m$.
In the paper, the approximate gradients for histogram binning are given as
\begin{align}
\frac{\partial \ndp}{\partial F(q)} & \approx \sum_{x\in \Xp}
\frac{\partial \delta(D(q,x),k)}{\partial D(q,x)}
\frac{\partial D(q,x)}{\partial F(q)},
\\
\frac{\partial \ndp}{\partial F(x)} & \approx \boldsymbol{1}[x\in\Xp]
\frac{\partial \delta(D(q,x),k)}{\partial D(q,x)}
\frac{\partial D(q,x)}{\partial F(x)},
\end{align}
where $q$ is a query patch and $S_q^+$ is the set of its matches in the database, and $D$ is the distance metric being learned.

In the real-valued case, the descriptor $F$ is modeled as a vector of neural network activations, with $L_2$ normalization:
\begin{align}
F_0(x)  =(f_1(x;w), f_2(x;w),\ldots, f_m(x;w))\in\mathbb{R}^m, ~~~  F(x) = \frac{F_0(x)}{\|F_0(x)\|}.
\end{align}
$D$ is now the Euclidean distance between unit vectors, whose partial derivative ${\partial D}/{\partial F}$ is
\begin{align}
\frac{\partial D(x,x')}{\partial F(x)} = \frac{\partial\sqrt{2-2F(x)^\top F(x')}}{\partial F(x)}
= \frac{-F(x')}{D(x,x')}.
\end{align}
Lastly, backpropagation through the $L_2$ normalization operation is as follows:
\begin{align}
\frac{\partial h_k^+}{\partial F_0(x)} =
\frac{1}{\|F_0(x)\|}\left[\frac{\partial h_k^+}{\partial F(x)} - F(x)\left(F(x)^\top\frac{\partial h_k^+}{\partial F(x)}\right)\right].
\end{align}

\section{Spatial Transformer Module}

We use the Spatial Transformer module in our networks to handle geometric noise and align input patches.
As is standard practice, the Spatial Transformer is initialized to output identity (directly copy input patches), and the learning rate of the affine transformation prediction layer is scaled down by 100x compared to other layers in the network.

A na\"ive application of the Spatial Transformer, however, leads to the boundary effect \cite{IC-STN}: when the predicted transformation requires sampling outside the boundaries of the input, the default zero-padding creates unfilled boundaries in the output. 
Since the input patches to the Spatial Transformer have limited size (42x42 in our network), out-of-boundary sampling frequently happens in operations such as zooming out and rotation, and can affect alignment by introducing unwanted image gradients.
Instead, we first pad the input patch by repeating its boundary pixels,\footnote{Implemented in Matlab using the \texttt{replicate} mode of the \texttt{padarray} function.} and then sample according to the predicted transformation, which prevents sharp gradients near boundaries. 
This is visually illustrated in Fig.~\ref{fig:st}, using patches from the challenging HPatches dataset, which has the largest amount of geometric noise among the datasets that we consider.
Although using zero padding still produces decent alignment, it affects the appearance of sampled patches, and does not help to improve final performance.
Our boundary padding produces much more visually plausible patches, and gives a good approximation to re-sampling from the original images. 

\begin{figure}
\centering
\includegraphics[width=7em]{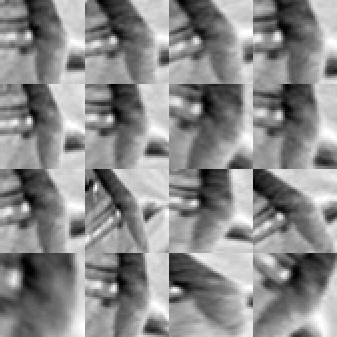}~~~
\includegraphics[width=7em]{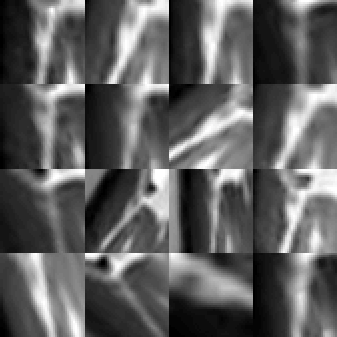}~~~
\includegraphics[width=7em]{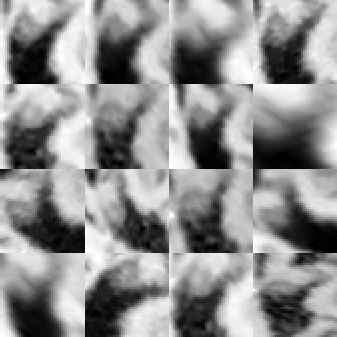}~~~
\includegraphics[width=7em]{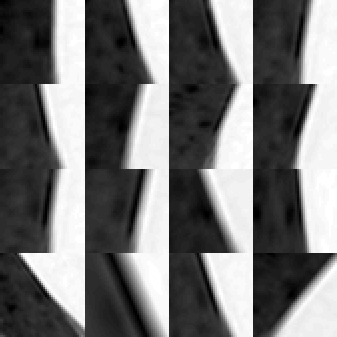}~~~
\includegraphics[width=7em]{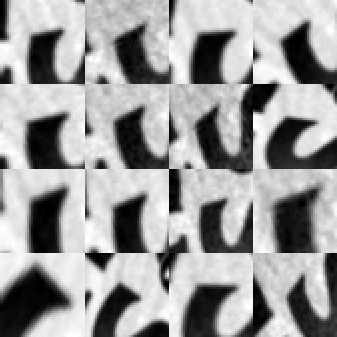}

\vspace{.5em}
\includegraphics[width=7em]{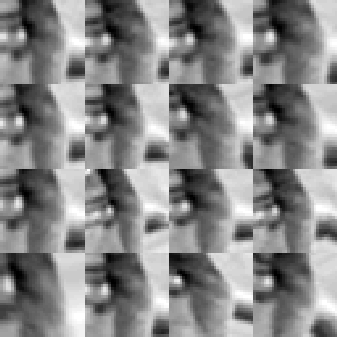}~~~
\includegraphics[width=7em]{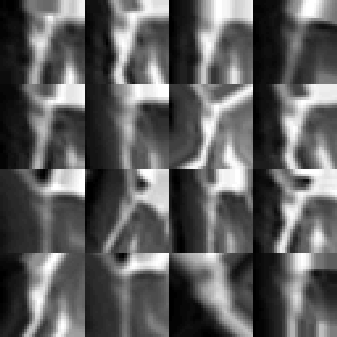}~~~
\includegraphics[width=7em]{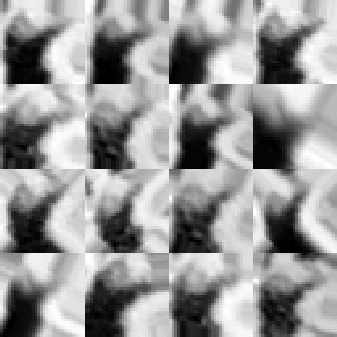}~~~
\includegraphics[width=7em]{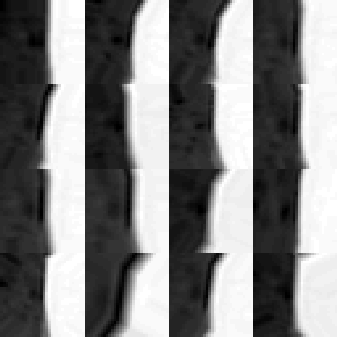}~~~
\includegraphics[width=7em]{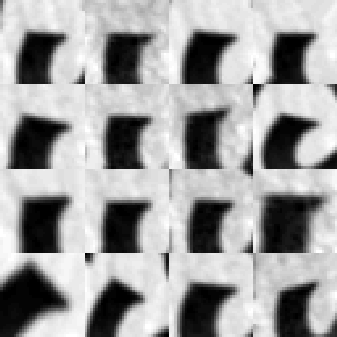}

\vspace{.5em}
\includegraphics[width=7em]{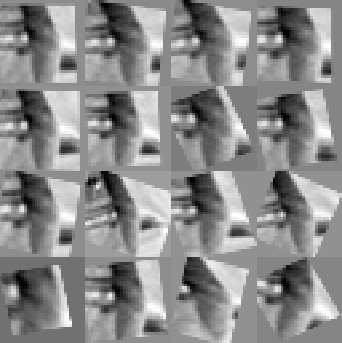}~~~
\includegraphics[width=7em]{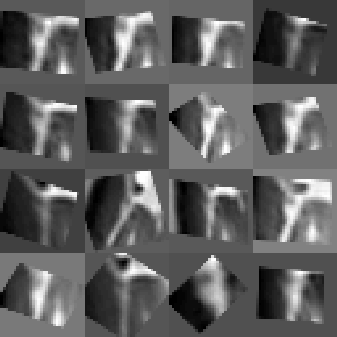}~~~
\includegraphics[width=7em]{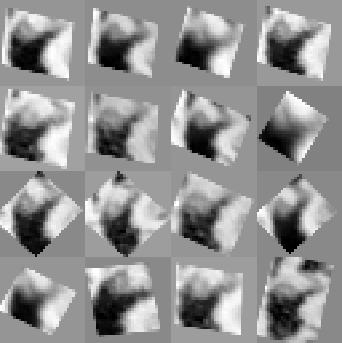}~~~
\includegraphics[width=7em]{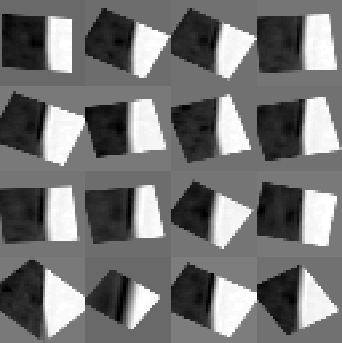}~~~
\includegraphics[width=7em]{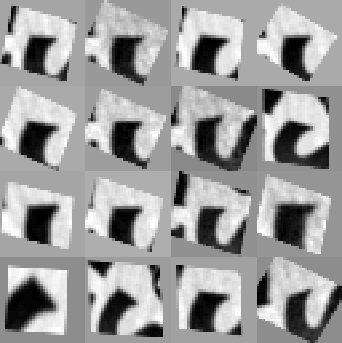}

\vspace{.5em}
\caption{Alignment using the Spatial Transformer in HPatches, where patches come in groups of 16.
The aligned patches are used as inputs to the descriptor network.
First row: original patches. Second row: aligned patches, using our boundary padding. Third row: aligned patches, using the default zero padding.
}
\label{fig:st}
\end{figure}

\section{Label Mining in HPatches}
As mentioned in Sec.~4.2, in the patch retrieval task in HPatches, the set of distractors for each query only consists of out-of-sequence patches. This differs from the image matching task where all distractors are in-sequence.
We use clustering to supply in-sequence distractors when optimizing patch retrieval performance.

\subsection{Clustering}
Since the 3D point correspondence for each training patch is given, it may appear that we can simply mark all patches that do not correspond to a certain 3D point as distractors for the corresponding patch.
However, the risk is that when an image has repeating structures (\eg windows on a building), patches that correspond to different 3D points could have nearly identical appearance, and forcing the network to distinguish between them would cause overfitting.
Instead, we need a mechanism to mark distractors only when the appearance difference is above a threshold.
Our solution is to use clustering: 
given an image sequence, we cluster all patches from this sequence by visual appearance.
Then, a threshold is put on the inter-cluster distances to determine distractors.

We use handcrafted visual features to represent patches in clustering. The best feature found in our experiments  is a combination of HOG \cite{HOG} and raw pixel values, which captures both the geometric and illumination patterns.
It is constructed as follows: a patch is resized to 64x64 to extract HOG features with 8x8 cell size, and then the same patch is resized to 16x16 and appended to the feature vector. The final feature dimensionality is 2240. 
Afterwards, we perform $K$-means clustering with $K=100$ clusters.
To derive a distance threshold, we compute all the pairwise distances between the cluster centers, and set the threshold at the $p$-th percentile of these distances.
If two clusters have larger distance than the threshold, their patches are considered distractors for each other. Otherwise, they are considered ``too visually similar'', and are ignored from each other's distractor set.
We use $p=20$.
Label mining is demonstrated in Fig.~\ref{fig:kmeans}.

\begin{figure}
\centering
\includegraphics[width=\linewidth]{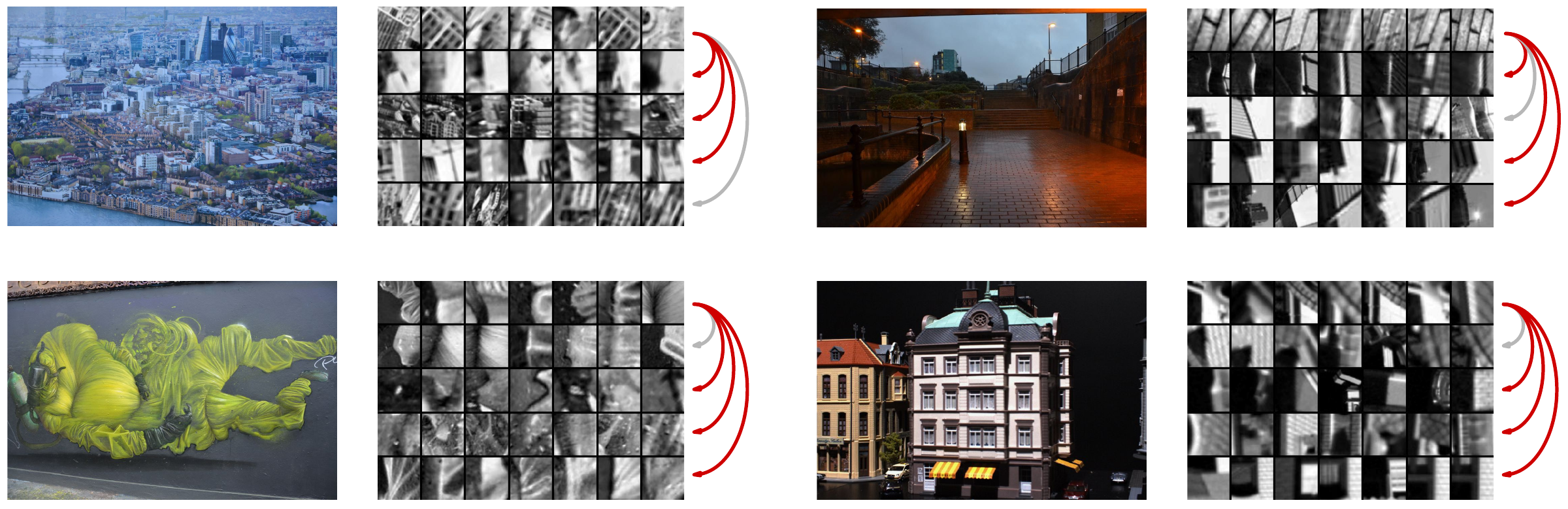}
\caption{
We demonstrate label mining in HPatches, using four randomly selected image sequences. First row: \texttt{v\_london}, \texttt{i\_steps}. Second row: \texttt{v\_maskedman}, \texttt{i\_yellowtent}.
The first image in each sequence is shown on the left, and on the right we visualize 5 randomly selected patch clusters, obtained using $K$-means.
Each row corresponds to a cluster.
A red arrow between clusters indicates that the inter-cluster distance is above a threshold, and their patches are used as distractors for each other. A gray arrow means that the inter-cluster distance is not high enough.
Patches are generally more similar in appearance within the same sequence  than across sequences, therefore mining the in-sequence distractors provides meaningful ``hard negatives'' for the learning.
}
\label{fig:kmeans}
\end{figure}

\subsection{Minibatch Sampling}
There are 76 image sequences in the training set of HPatches.
Without label mining, we uniformly sample patch groups from all sequences to construct training minibatches, so on average only about 1/76 of the patches in each minibatch are from the same sequence.
In this case, even if the in-sequence distractor labels are known, their contribution to the gradients is limited.
Therefore,  we use a modified minibatch sampling strategy when label mining is in effect, so that more patches from the same sequence are placed in a minibatch. 

Specifically, to construct a minibatch, we first sample two image sequences. 
Then, an equal number of patch groups (each containing 16 matching patches) are sampled from each sequence. For example, if batch size $M=1024=64\times 16$, then 32 groups are sampled from each of the two sequences. 
This way, for each patch, roughly half of its distractors are out-of-sequence patches, and the other half are in-sequence, which are generally harder to distinguish.
This simple heuristic gave about 6\% absolute improvement in image matching mAP in our experiments, and we did not specifically tune the ratio of in-sequence \vs out-of-sequence distractors.
With this strategy, a minibatch involves a pair of sequences, and a training epoch loops over all the $\frac{76\times(76-1)}{2}=2850$ pairs, and takes less than 10 minutes with $M=1024$ in our GPU implementation.

\section{Experimental Details}
We train our networks from scratch using SGD.
The initialization scheme proposed in  \cite{heinit} is adopted, since the architecture uses ReLU activations.
Through validation experiments, we found that an initial learning rate of 0.1 works well with batch size $M=1024$ in all datasets. 
For other batch sizes, we scale the learning rate linearly, according to the suggestion in \cite{imagenet1hr}.
For UBC Phototour,  inspired by HardNet \cite{HardNet}, the learning rate is decreased linearly to zero within 100 epochs.
For HPatches, we actually found a more traditional strategy to work better: we use a constant learning rate and divide it by 10 every 10 epochs, for 32 epochs total.

For RomePatches, the training set has 10,000 patches, or 1,000 groups of 10 patches, which is quite small. 
To stabilize the training, we increase the number of minibatches in each epoch to 1,000 as follows:  the $k$-th batch first includes the $k$-th group, and then randomly samples other groups to fill the batch.
With this strategy, each epoch processes the training set multiple times, and we found 5 epochs to be sufficient to ensure convergence.

Our implementation uses MatConvNet \cite{MatConvNet}.
For competing methods, we use the publicly released models/implementations.
\begin{itemize}
\vspace{-.2em}
\item We use pretrained L2Net models\footnote{\url{https://github.com/yuruntian/L2-Net}}. We use the versions trained with data augmentation.
\vspace{-.2em}
\item We use pretrained HardNet models\footnote{\url{https://github.com/DagnyT/hardnet}}. We use the versions trained with data augmentation.
\vspace{-.2em}
\item For SIFT and LIOP, we use the implementation in VLFeat \cite{vlfeat}.
\end{itemize}

Performance on HPatches is evaluated using the HPatches benchmark\footnote{\url{https://github.com/hpatches/hpatches-benchmark}}.
For the image matching experiment in Oxford dataset,  the detection of interest points and extraction of patches are performed using the \texttt{vl\_covdet} function in VLFeat, with the \texttt{PatchRelativeExtent} parameter set to 3.

\end{document}